\documentclass[10pt,twocolumn,letterpaper]{article}

\usepackage[margin=.65in]{geometry}

\usepackage{times}
\usepackage{epsfig}
\usepackage{graphicx}
\usepackage{amsmath}
\usepackage{amssymb}
\usepackage{epstopdf}
\usepackage{float}
\usepackage{multirow}

\usepackage[margin=15pt,font=small,labelfont=bf]{caption}

\date{ }

\begin{document}

\title{Hi Detector, What's Wrong with that Object?
Identifying Irregular Object From Images by Modelling the Detection Score Distribution}

\author{Peng Wang$^1$\thanks{This work was done when P. Wang was visiting The
  University of Adelaide.},
  Lingqiao Liu$^2$, Chunhua Shen$^2$\thanks{E-mail: chunhua.shen@adelaide.edu.au}, Anton van den Hengel$^2$, Heng Tao Shen$^1$\\
$^1$ The University of Queensland, Australia;
~
~
~
$^2$ The University of Adelaide, Australia
}

\maketitle

\begin{abstract}

  In this work, we study the challenging problem of identifying the irregular status of objects from images in an ``open world'' setting,
  that is, distinguishing the irregular status of an object category from its regular status as well as objects from other categories in the absence of ``irregular object'' training data. To address this problem, we propose a novel approach by inspecting the distribution of the detection scores at multiple image regions based on the detector trained from the ``regular object'' and ``other objects''. The key observation motivating our approach is that for ``regular object'' images as well as ``other objects'' images, the region-level scores follow their own essential patterns in terms of both the score values and the spatial distributions while the detection scores obtained from an ``irregular object'' image tend to break these patterns. To model this distribution, we propose to use Gaussian Processes (GP) to construct two separate generative models for the case of the ``regular object'' and the ``other objects''. More specifically, we design a new covariance function to simultaneously model the detection score at a single region and the score dependencies at multiple regions.
We finally demonstrate the superior performance of our method on a large dataset newly proposed in this paper.
\end{abstract}

\begin{figure*}[t!]
\begin{center}
\includegraphics[scale=.427]{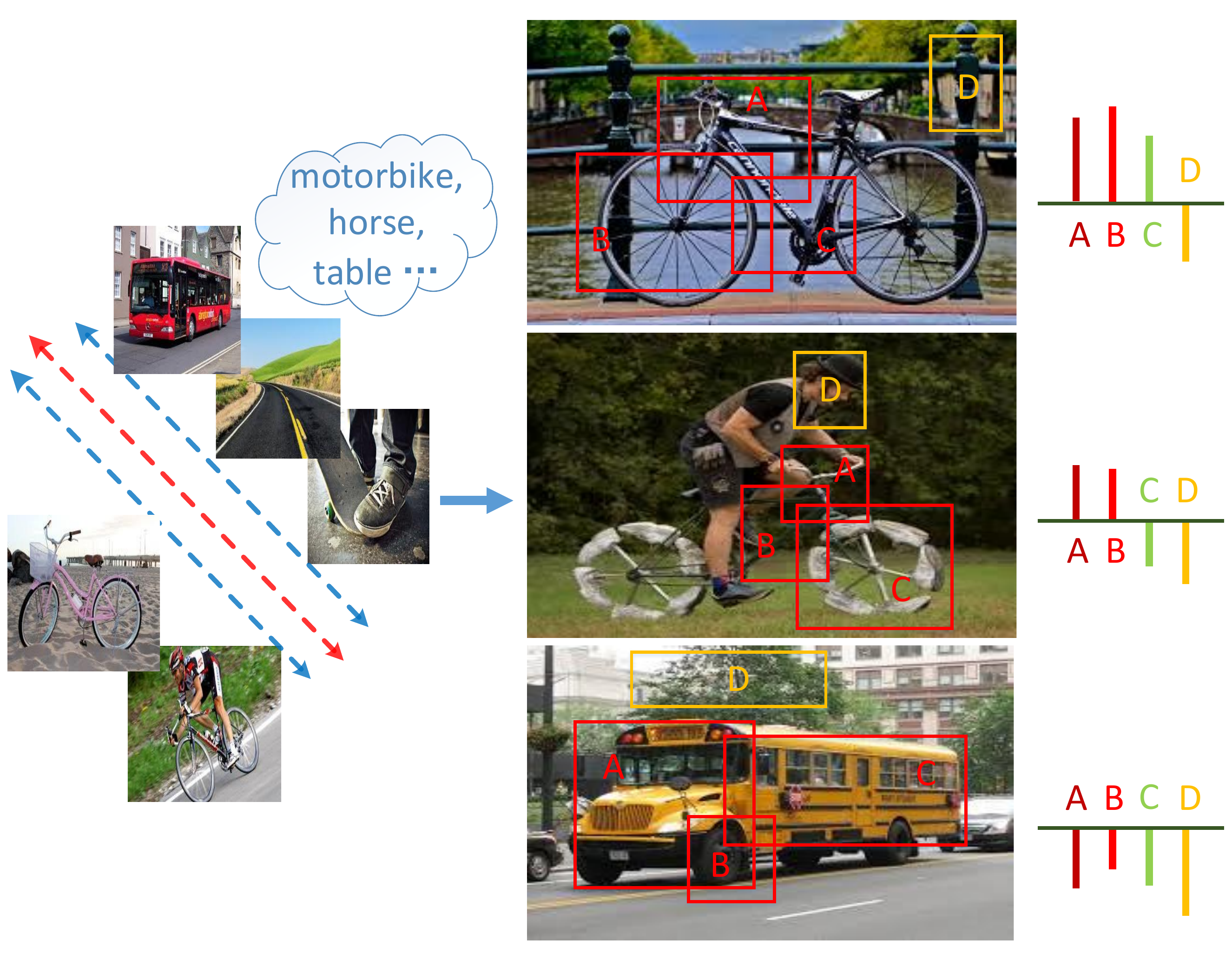}
\end{center}
   \caption{Illustration of our idea for detecting the irregular ``bicycle''. By applying a detector learned from the ``regular bicycle'' and ``other objects'' to multiple image regions, we classify the ``regular bicycle'', ``irregular bicycle'' and ``other object'' (``bus'' in this figure) through the distribution of the detection scores. The discriminative information lies in both the values of the detection scores and the spatial dependency patterns of those scores, e.g., the score dependency between neighbouring proposals B and C.}
\label{fig:intro_fig}
\end{figure*}

\tableofcontents\newpage

\section{Introduction}

Humans have the ability to detect the irregular status of objects without seeing the irregular patterns beforehand. Mimicking this ability with computer vision technique can be practically useful for the applications such as surveillance or quality control. Existing studies towards this goal are usually conducted on small datasets and controlled scenarios i.e., with relatively simple background \cite{Boiman:2007} or specific type of irregularity \cite{Choi2012853,Park:2012}. To address this issue, in this work we present a large dataset which captures more general irregularities and has more complex background. Moreover, we adopt a more realistic ``open world'' evaluation protocol. That is, we need to distinguish the ``irregular version of object-of-interest'' not only from the ``regular object'' belonging to the same category but also from the ``other objects'' (objects from other categories) at the test stage.

The reason why people can recognize an image to be an irregular example of a certain object is because it shares some common patterns of this object but deviates from the regular examples of the object. In other words, the ``irregular object'' images are supposed to be more similar to the ``regular object'' images comparing to the images of other objects. If we apply a detector learned using ``regular object'' images as positive training data and images of other objects as negative data to the regions of an image, the score values of the ``irregular object'' images are expected to be larger than the scores of the ``other object'' images and smaller than those of the ``regular object'' images. Apart from the values of the region-level detection scores, the spatial distributions of the detection scores may encode some discriminative information as well. As illustrated in Fig.~\ref{fig:intro_fig}, positive detection scores should be densely overlapped in regular images while in irregular images the score distribution may break this pattern due to the existence of the irregular parts. To model these two factors, we propose to use Gaussian Processes (GP) \cite{Rasmussen:2005} to construct two separate generative models for the detection scores of ``regular object'' image regions and ``other objects'' image regions. The mean function is defined to depict the prior information of the score values of either ``regular object'' images or ``other object'' images and a new covariance function is designed to simultaneously model the detection score at a single region non-parametrically and capture the inter-dependency of scores at multiple regions. Note that unlike the conventional use of GP in computer vision, our model does not assume that the region scores of an image are i.i.d. This treatment allows our method to capture the spatial dependency of detection scores,
which turns out to be crucial for identifying irregular objects.
By comparing with several alternative solutions on the proposed dataset, we experimentally demonstrate the effectiveness of the proposed method. To summarize, the main contributions of this paper are:

\begin{itemize}
\item We propose a large dataset and present a more realistic ``open world'' evaluation protocol for the task of irregular object identification from images.

\item We propose a novel approach for irregularity detection by looking into the detection score values as well as the spatial distributions of the detection scores of the image regions. We propose to use Gaussian Processes (GP) to simultaneously model the detection score at a single region and the score dependencies at multiple regions.

\end{itemize}

\section{Related Work}

\noindent\textbf{Irregular Image/Video Detection.} There exists a variety of work focusing on irregular image and/or video detection. While some approaches attempt to detect irregular image parts or video segments given a regular database \cite{BinZhao:2011,Boiman:2007,1315249,Hamid05detectionand}, other efforts are dedicated to addressing some specific types of irregularities \cite{Park:2012, Choi2012853} such as out-of-context via building some corresponding models.

Standard approaches for irregularity detection are based on the idea of evaluating the dissimilarity from regular. The authors of \cite{1315249,Hamid05detectionand} formulate the problem of unusual activity detection in video into a clustering problem where unusual activities are identified as the clusters with low inter-cluster similarity. The work \cite{Boiman:2007} detects the irregularities in image or video by checking whether the image regions or video segments can be composed using large continuous chunks of data from the regular database. Despite the good performance in irregularity detection, this method severely suffers from the scalability issue, because it requires to traverse the database given any new query data. Sparse coding \cite{Lee07efficientsparse} is employed in \cite{BinZhao:2011} for unusual events detection. This work is based on the assumption that unusual events cannot be well reconstructed by a set of bases learned from usual events. It is claimed in \cite{BinZhao:2011} that it has advantages comparing to previous approaches in that it is built upon a rigorous statistical principle.

Another stream of work focus on addressing specific types of irregularities. The work of \cite{5540221, Choi2012853} focus on exploiting contextual information for object recognition or out-of-context detection, like ``car floating in the sky''. In \cite{5540221}, they use a tree model to learn dependencies among object categories and in \cite{Choi2012853} they extend it by integrating different sources of contextual information into a graph model. The work \cite{Park:2012} focuses on finding abnormal objects in given scenes. They consider wider range of irregular objects like those violate co-occurrence with surrounding objects or violate expected scale. However, the applications of these methods are very limited since they rely on pre-learned object detector to accurately localize the object-of-interest.

\noindent\textbf{Gaussian Processes in Computer Vision.} Due to the advantage in nonparametric data fitting, GP has widely been used in the fields like classification \cite{Altun:2004}, tracking \cite{1640765}, motion analysis \cite{Kim11gaussianprocess} and object detection \cite{Vezhnevets14,Vezhnevets15}. The work \cite{Kim11gaussianprocess} uses GP regression to build spatio-temporal flow to model the motion trajectories for trajectory matching. In \cite{Vezhnevets14,Vezhnevets15}, object localization is done via using GP regression to predict the overlaps between image windows and the ground-truth objects from the window-level representations.
\section{A New Dataset}
\begin{figure*}
\begin{center}
\includegraphics[scale=.526]{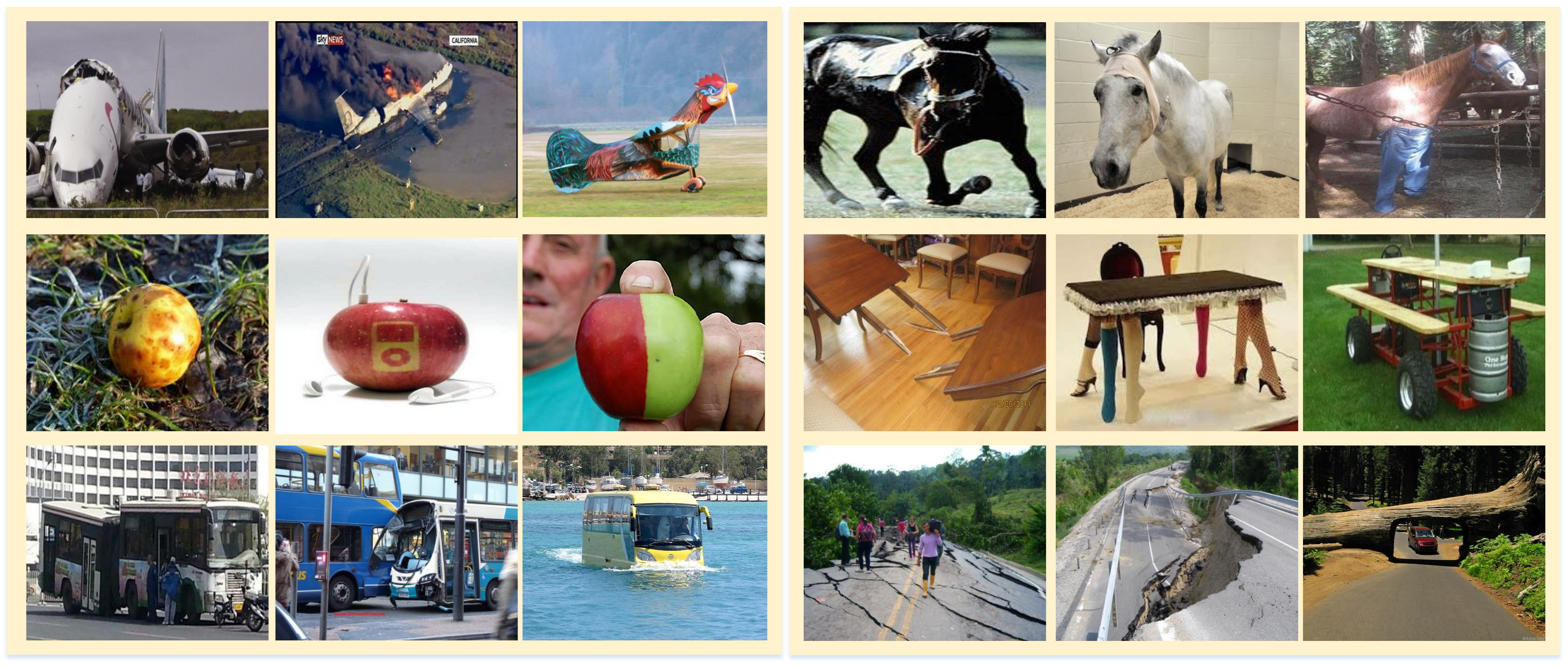}
\end{center}
   \caption{Examples of irregular images. Left column: aeroplane, apple, bus; Right column: horse, dining table, road.}
\label{fig:sampled_frms}
\end{figure*}

\subsection{Dataset Introduction}

In this paper, we propose a new dataset for the task of irregular image detection. The data is collected from \textit{Google Images} and \textit{Bing Images} which is composed of 20,420 images belonging to 20 classes. We choose the 20 classes referring to the PASCAL VOC dataset \cite{Everingham:2010} but replace some classes that are not suitable for the task. For example, it is hard to define ``irregular person''. The images of each class are composed of both regular images and irregular images. For regular images, we try different feasible queries to collect sufficient data. Taking ``apple'' for example, we try ``fuji apple'', ``pink lady'', ``golden delicious'', etc. To collect irregular images, we use keywords like ``irregular'', ``unusual'',  ``abnormal'', ``weird'', ``broken'', ``decayed'', ``rare'', etc. After the images are returned, we manually remove the unrelated and low-quality data. Also, we perform near-duplicate detection to remove some duplicate images.\footnote{This dataset will be released to facilitate further research.} Fig.~\ref{fig:sampled_frms} shows some examples of irregular images.

\begin{table}[h]
\caption{Comparison of the proposed dataset to existing datasets. \cite{Choi2012853} addresses the irregular type of \textbf{out of context}. \cite{Park:2012} deals with violations of \textbf{co-occurrence, positional relationship} and \textbf{scale.}}
  \centering

        \begin{tabular}{cccc}

            \hline\noalign{\smallskip}

                dataset  &    $\#$ images  & irregular category & accurate detector \\

            \noalign{\smallskip}

            \hline

            \noalign{\smallskip}

             \cite{Park:2012}   & 150  &  specific & yes\\

             \cite{Choi2012853} & 218  &	 specific & yes\\
			 ours  & 20,420 & general & no\\

            \noalign{\smallskip}

            \hline \noalign{\smallskip}

      \end{tabular}

  \label{tab:dataset}%
\end{table}%

There exists some other datasets \cite{Choi2012853,Park:2012} for irregular image detection. A comparison between our dataset and the existing datasets is summarized in Table \ref{tab:dataset}. The main difference is twofold.
\begin{itemize}
\item Our dataset is large-scale comparing to the existing datasets, increasing the number of images from several hundred to more than twenty thousand.
\item While the existing datasets are proposed for specific irregular category such as ``out-of-context", ``relative position violation" and ``relative scale violation", our dataset is for general irregular cases.

\end{itemize}
Besides the above differences, we adopt a more practical evaluation protocol compared with \cite{Choi2012853,Park:2012}. That is, we evaluate the irregular object detection with the presence of irrelevant objects. This is different from \cite{Boiman:2007} where irregularity detection is performed in controlled environment with relatively simple background.

\subsection{Problem Definition}
For a given object category $\mathcal{C}$, we divide it into two disjoint subcategories, a regular sub-class $\mathcal{C}^r$ and an irregular sub-class $\mathcal{C}^u$, with $\mathcal{C} = \mathcal{C}^r \cup \mathcal{C}^u$ and $\mathcal{C}^r \cap \mathcal{C}^u = \emptyset$. We call an image $I$ a regular image if $I\in\mathcal{C}^r$ and an irregular image if $I\in\mathcal{C}^u$. If an image $I$ does not contain the given object, we label it as belonging to the ``other class'' set $\mathcal{C}^o$. The task is to determine if a test image $I\in\mathcal{C}^u$. Note that for $\mathcal{C}$, only the regular and ``other class'' images are available for training.

%
%
%
%
%
%
%
%
\section{Key Motivation}
Regular object images of the same class are alike; each irregular object image, however, is irregular in its own way. Thus, it is somehow impossible to collect a dataset to cover the space of the irregular images and one common idea to handle this difficulty is to build a ``regular object'' model to identify the ``irregular objects'' as outliers. While most traditional methods \cite{BinZhao:2011,Boiman:2007} build this model based on the visual features extracted from images, our approach takes an alternative methodology by firstly training a detector from the ``regular object'' images and ``other objects'' images and then discovering the irregularity based on the detection score patterns. The merit of using detection scores for irregularity detection are as follows. (1) It is more computationally efficient since the appearance information has been compressed to a single scalar of detection values. This enables us to explore complex interaction of multiple regions within an image while maintaining reasonable computational cost. (2) It naturally handles the background and ``other class'' distraction since our detector is trained by using the ``regular object'' and ``other objects''. More specifically, our method is inspired by two intuitive postulates of how humans recognize an ``irregular object", which are elaborated as follows.

\begin{figure}[t]
\begin{center}
\includegraphics[scale=.32]{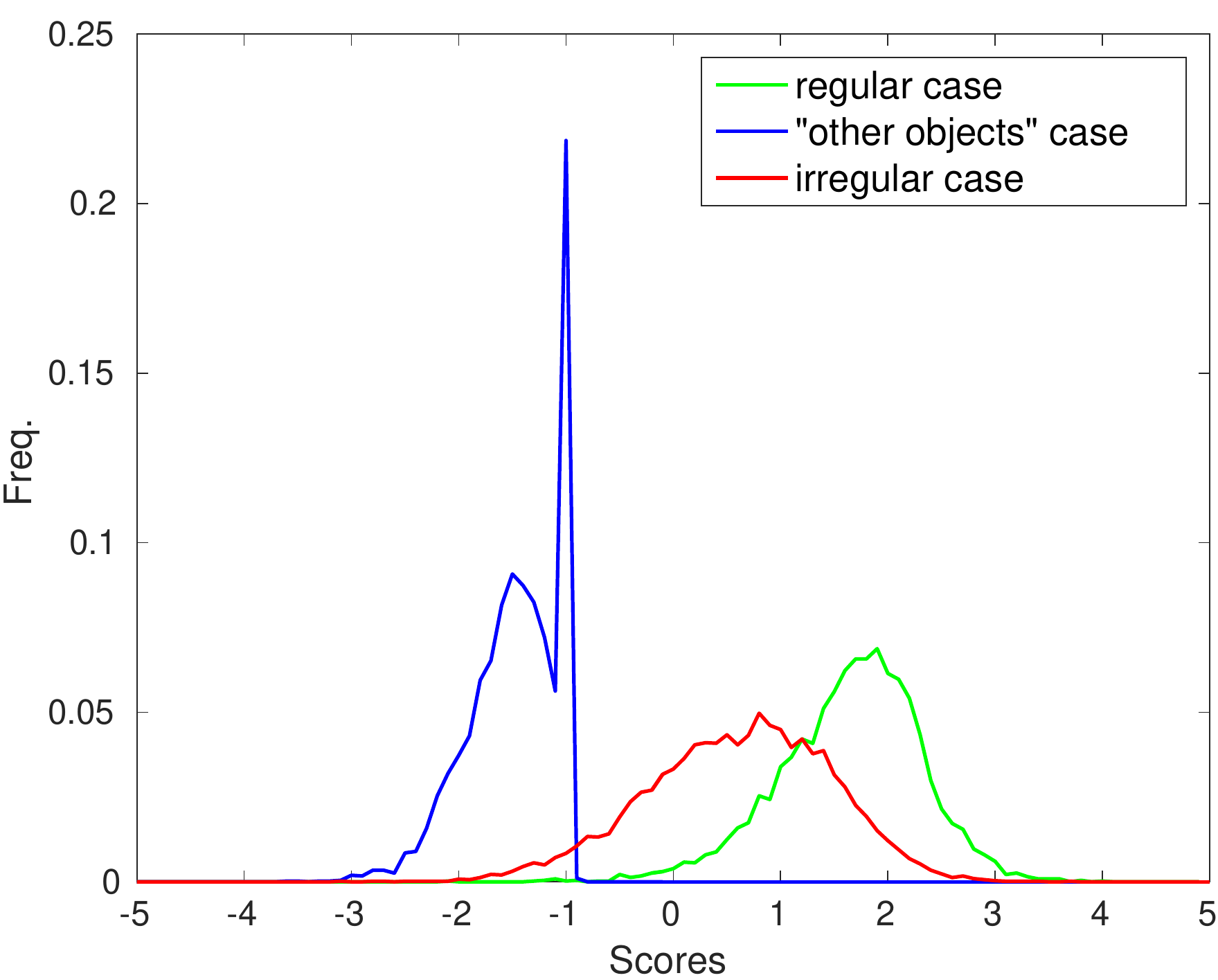}
\end{center}
   \caption{Histograms of decision scores for regular images, irregular images and ``other class'' images in the testing data. The decision scores are obtained by applying the classifiers learned from global images.}
\label{fig:hist}
\end{figure}

\noindent\textbf{Postulate I: discrimination in detection score values.} From the perspective of human vision, an irregular object is something ``looks like an object-of-interest, but is still different from its common appearance". If we view the object detection score as a measure of the likelihood of an image containing the object, then the above postulate could correspond to a relationship in detection scores $f(I^o)<f(I^u)<f(I^r)$, where $f(I^o)$, $f(I^u)$ and $f(I^r)$ denote the detection score of the ``other object'', ``irregular object'' and ``regular object'' respectively. To verify this relationship, we train an image-level object classifier and plot the accumulated histograms of the scores of regular, irregular and other-class images of each class in Fig.~\ref{fig:hist}. It can be seen from this figure that the distribution of the score values is generally consistent with our assumption. However, there are still overlaps especially between regular and irregular images, which means that using this criterion alone cannot perfectly distinguish the irregular images.

\begin{figure*}[h!]
\begin{center}
\includegraphics[scale=.524]{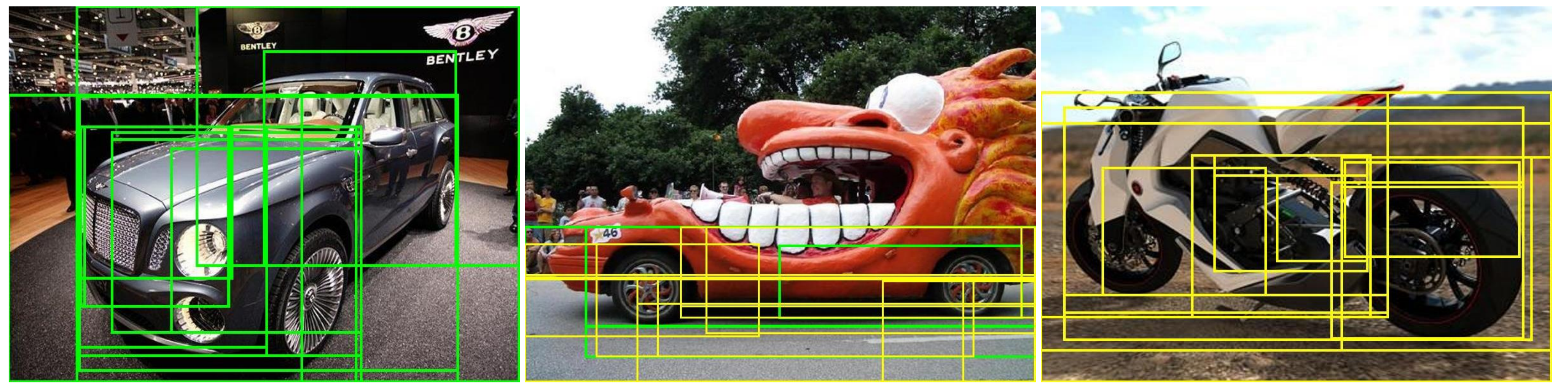}
\end{center}
   \caption{Visualization of spatial distribution of detection scores for test images of \textsl{car} class. Top-20 scored bounding boxes of an image are visualized. Positive proposals are visualized in green box and negative are visualized in yellow. From left to right: regular car, irregular car and other object (motorbike).} %
\label{fig:score_dis}
\end{figure*}

\noindent\textbf{Postulate II: discrimination in the spatial dependency of detection scores.} When exposed to part of the regular object, human can predict what the neighbouring parts of the object should look like without any difficulty. But irregular object may break this smoothness. This suggests that if we apply an object detector to the object proposals of an image, the region-level detection scores of the three different types of images may exhibit different dependency patterns. Fig.~\ref{fig:score_dis} shows the top 20 regions of some example images of \textsl{car} class according to the values of the detection scores. As seen, for regular car the positive bounding boxes are densely overlapped and images from other classes such as \textsl{motorbike} are supposed to have no positively scored proposals. Detection scores of irregular images may disobey both of these two distribution patterns. For example, two strongly overlapped regions may have opposite detection scores.

%

\label{sec:motivation}

\section{Proposed Approach}
Motivated by the above analysis, we propose a two-step approach to the task of irregular image detection. We first apply a Multi-Instance Learning (MIL) approach to learn a region-level object detector and then design Gaussian Processes (GP) based generative models to model the detection score distributions of the ``regular object'' and the ``other objects''. Once the model parameters are learned, we can readily determine whether a test image is irregular by evaluating its fitting possibilities to these two generative models.

\subsection{Object Detector Learning}
Taking the region proposals of images as instances, we represent each image as a bag of instances. Since we only have the image-level label indicating the presence or absence of the object, the learning of region-level detector is essentially a weakly supervised object localization problem. Considering both the localization accuracy and the scalability, we follow the MIL method in \cite{Oquab_2015_CVPR} to learn an object detector for each class. For a class $C$, we have a set of regular images containing the object as positive training data and a set of images belonging to other classes where the object concerned do not appear as negative training data.

We use Selective Search \cite{Uijlings13} to extract a set of object proposals for each image and from the perspective of MIL, each proposal is regarded as an instance. Then each image $I^i$ is represented by a $N_i\times{D}$ matrix $\mathbf{X}^i$ where $N_i$ denotes the number of proposals and $D$ represents the dimensionality of the proposal representations. Inspired by \cite{Oquab_2015_CVPR}, we optimize the following objective function to learn the detector,
\begin{align}
 \mathcal{J} = \sum_i{\mathrm{log}(1+e^{-y^i\max_{j}\{\mathbf{\mathbf{w}^T x}^i_{j} + b\}})},
\end{align}
where $\mathbf{w}\in\mathbb{R}^{D\times{1}}$ serves as an object detector, $\mathbf{x}^i_{j}$ indicates the $j$th instance of the $i$th image and $\mathbf{w}^T \mathbf{x}^i_{j} + b$ is its detection score. The single image-level score is aggregated via the max-pooling operator $\max\{\cdot\}$ and it should be consistent with the image-level class label $y^i\in\{1,-1\}$. The parameters $\mathbf{w}$ and $b$ can be learned via back-propagation using stochastic gradient descent (SGD).
\subsection{Gaussian Processes based Generative Models}
In this section, we elaborate how to use GP to model the distribution of the region-level detection scores. Unlike traditional GP based regression \cite{Vezhnevets15} which takes a single feature vector as input, we treat multiple proposals within an image as the input and our model will return a probability to indicate the fitting likelihood of the proposal set.

GP assumes that any finite number of random variables drawn from the GP follow a joint Gaussian distribution and this distribution is fully characterized by a mean function $m(x)$ and a covariance function $k(x,x')$ \cite{Rasmussen:2005}. In our case, we treat the detection score of each proposal as a random variable. The mean function depicts the prior information of the score values, e.g. the value tends to be a positive scalar for the ``regular object'' images. The covariance function plays two roles. (1) As in standard GP regression, it serves as a non-parametric estimator of the score value. More specifically, if a proposal is similar (in terms of a defined proposal representation) to a proposal in the training set, it encourages them to share similar scores. (2) As one of our contributions, we also add a term in the covariance function to encourage the overlapped object proposals within the same test image to share similar detection scores.
In the following subsections, we introduce the details of the design of the mean function and covariance function.
\subsubsection{GP Construction}
For each class $\mathcal{C}$, we will construct two GP based generative models for regular images and ``other objects'' images separately. Without losing generality, we will focus on regular images in the following part.

Suppose we have $N^\mathcal{C}$ positive training images for class $\mathcal{C}$. For each image $I^i$ ($i\in\{1,2,\cdots,N^\mathcal{C}\}$), we use the top-$n$ scored proposals $s^i_{j} $ ($j\in\{1,2,\cdots,n\}$) only in order to reduce the distraction impact of the background. Their associated detection scores can be obtained via the function $f(s^i_j)$. In our model we assume that $f$ is distributed as a GP with a mean function $m(\cdot)$ and a covariance function $k(\cdot,\cdot)$
\begin{align}
f \sim \mathcal{GP}(m, k).
\end{align}
\noindent\textbf{Mean function:} We define the mean function $m(s) = \mu$, where $\mu$ is a scalar constant learned through parameter estimation. It can be intuitively understood as the bias of the detection score in the regular object or other object cases. For example, it tends to be a positive (negative) value for the ``regular (other) object'' case.

\noindent\textbf{Covariance function:} As aforementioned analysis, the covariance function is decomposed into two parts, an inter-image part and an inner-image part. While the inter-image part is employed to regress the proposal-level detection score in the light of the proposals in the training set, the inner-image part is used to model the dependencies of the scores within one test image. To define the inter-image covariance function for a proposal pair belonging to different images, it needs to design a representation for each proposal so that their similarity can be readily measured. We leverage the spatial relationship between a proposal and the proposal with the maximum detection score within the same image as this representation.
More specifically, assuming the maximum-scored proposal in an image $I^i$ is $s^i_{max}$, the representation of a proposal $s$ in $I^i$ is defined as,
\begin{align}
\phi(s) = [\mathrm{IoU}(s,s^i_{max}), c(s,s^i_{max})],
\label{eq:inter-cov}
\end{align}
where $\mathrm{IoU}(s,s^i_{max})$ denotes the intersection-over-union between $s$ and $s^i_{max}$ and $c(s,s^i_{max})$ denotes the normalized distances between the centers of $s$ and $s^i_{max}$. Note that these two measurements reflect a proposal's overlapping degree, distance to the maximum-scored proposal and indirectly the size of the proposal. Intuitively, these factors could be used to predict the detection score value of a proposal.

With this representation, we can define the inter-image covariance function $k_{inter}(s,s')$ of $s$ and $s'$ as,
\begin{align}
 \mathrm{exp}\bigg(-\frac{1}{2}\big(\phi(s)-\phi(s')\big)^T\mathrm{diag}(\mathbf{\gamma})\big(\phi(s)-\phi(s')\big)\bigg),
\end{align}
where $\mathrm{diag}(\mathbf{\gamma})$ is a diagonal weighting matrix to be learned.

The inner-image covariance function serves as one of the key contributions of this work, which poses a smoothness constraint over the scores of the overlapped object proposals in an image. For a pair of inner-image proposals $s$ and $s'$, we define the inner-image covariance function as follows \footnote{If two proposals $s$ and $s'$ are from different images, $k_{inner}(s,s')=0$},
\begin{align}
k_{inner}(s,s')=
\frac{2S(s\cap{s'})}{S(s\cap{s'})+S(s\cup{s'})},
\end{align}
where $S$ stands for the area. Note that the formula is variant to standard intersection-over-union \cite{Everingham:2010} commonly used as detection metric. The reason why we define it like this is because it is exactly $\chi^2$ kernel and can guarantee the covariance matrix to be positive definite \cite{10.1109/TPAMI.2011.153}.

With both the inter-image and inner-image covariance function, we can obtain the overall covariance function of any proposal pair $s$ and $s'$ as,
\begin{align}
k(s,s') = a\cdot{k_{inner}}(s,s')+b\cdot{k_{inter}}(s,s'), \label{Equ:conv}
\end{align}
where $a,b$ are hyper-parameters regulating the weights of these two kernel functions.

\subsubsection{Hyper-parameter Estimation}
In this part, we introduce the hyper-parameter learning for the GPs. Still, we use regular images for description. In the definition of the mean and covariance functions of the GP, we introduce the hyper-parameters $\theta = \{\mu, \gamma, a, b\}$. We estimate the hyper-parameters by minimizing the negative logarithm of the marginal likelihood of all the detection scores of the training proposals given the hyper-parameters,
\begin{align}
-L = -\mathrm{log}~p(f(\mathcal{S})|\mathcal{S},\theta),
\end{align}where $\mathcal{S}$ denotes the training proposals and $f(\mathcal{S})$ denotes their detection scores. We use the toolbox introduced in \cite{Rasmussen:2010:GPM:1756006.1953029} for hyper-parameter optimization.

\subsubsection{Test Image Evaluation}
For class $\mathcal{C}$, let $\mathbf{s_r}$ be a set of proposals of regular training images and $\mathbf{f_r}$ be their detection scores. We can establish the covariance matrix $K$ for the training data. Given a target set of proposals $\mathbf{s_t}$ from a test image and their detection scores $\mathbf{f_t}$, the joint distribution of $\mathbf{f_r}, \mathbf{f_t}$ can be written as,

\begin{align}
\begin{bmatrix}
\mathbf{f_r} \\
\mathbf{f_t}
\end{bmatrix}
\sim \mathcal{N}\Big(
\begin{bmatrix}
\mathbf{\mu} \\
\mathbf{\mu}
\end{bmatrix},
\begin{bmatrix}
K & k(\mathbf{s_r, s_t}) \\
k(\mathbf{s_r, s_t})^T & k(\mathbf{s_t, s_t})
\end{bmatrix}
\Big),
\end{align}
where $\mathbf{\mu}$ is the mean vector, $k(\mathbf{s_r, s_t})$ calculates the inter-image covariance matrix between training set and testing set and $k(\mathbf{s_t, s_t})$ calculates the inner-image covariance of the test data. The fitting likelihood of the testing set to the generative model of the regular images can be expressed as,
\begin{align}
\begin{split}
\mathbf{f_t}|\mathbf{f_r} \sim \mathcal{N}\bigg(\mathbf{\mu}+k(\mathbf{s_r,s_t})^TK^{-1}(\mathbf{f_r-\mu}), \\
k(\mathbf{s_t,s_t})-k(\mathbf{s_r,s_t})^TK^{-1}k(\mathbf{s_r,s_t})\bigg).
\end{split}
\end{align}
Similarly, we can obtain the likelihood of the testing set given the ``other class'' training set. After obtaining the likelihood of the testing set given both regular training data and ``other class'' training data, we can compute the logarithm of the overall fitting likelihood of $\mathbf{f_t}$ as
\begin{align}\label{Eq:final_decision}
\mathrm{max}\big(\mathrm{log}~p(\mathbf{f_t}|\mathbf{f_r}), \mathrm{log}~p(\mathbf{f_t}|\mathbf{f_o})\big),
\end{align}
where $\mathbf{f_o}$ represents the scores of ``other class'' training set. For either regular or ``other class'' test images, they could fit one of the generative models better than the irregular images. In other words, irregular images are supposed to obtain lower values in Eq. (\ref{Eq:final_decision}).
\section{Experiments}

\subsection{Experimental Settings}
In this paper, we use the pre-trained CNN model \cite{Simonyan15} as feature extractors for object detector learning. Specifically, we use the activations of both the second fully-connected layer and the last convolutional layer as the representation of the object proposal or the whole image. Feeding an image into the CNN model, the activations of a convolutional layer are $n\times{m}\times{d}$ (e.g., $14\times{14}\times{512}$ for the last convolutional layer) with $n,m$ corresponding to different spatial locations and $d$ the number of feature maps. Given a proposal, we aggregate the convolutional features covered by it via max pooling to obtain the proposal-level convolutional features. We perform $L2$ normalization to these two types of features separately and concatenate them as the final representation. The dimensionality of the features is 4,608.

For each class, we construct GP based generative models for regular images and ``other class'' images separately. For regular images, we initialize the value of the mean function as $3$ and for ``other class'' images we set the initial value to be $-3$. The hyper-parameters $a, b$ in Eq.~(\ref{Equ:conv}) are both initialized to be 0.5 and $\gamma$ is initialized randomly. We use the top-20 scored proposals of each image for both generative model construction and test image evaluation. The test data of each class is divided into three parts including regular images, irregular images and images belonging to other classes. We label irregular images as $1$ and label regular and ``other class'' images as $-1$. Mean Average Precision (mAP) is employed to evaluate the performances of the approaches.

\subsection{Experimental Results}
\subsubsection{Alternative Solutions}
\begin{table*}[ht]\footnotesize
\caption{Experimental results. Average precision for each class and mAP are reported.}
  \centering

        \begin{tabular}{cccccccccccc}

        \hline\noalign{\smallskip}

Methods & aeroplane & apple & bicycle & boat & building & bus & car & chair & cow & dinging table \\

            \noalign{\smallskip}

            \hline

            \noalign{\smallskip}
			 Positive-negative Ratio & 58.0 & 26.6 & 50.4 & 52.4 & 60.0 & 37.8 & 55.4 & 48.7 & 31.6 & 28.8 \\
             Global SVM & 88.8 & 70.8 & 81.3 & 82.9 & 85.5 & 76.4 &  87.6 & 69.7 & 61.7 & 79.8\\
             MIL + Max & 86.9 &	70.0 & 85.0 & 78.8 & 81.7 & 77.6 & 87.8 & 70.5 & 63.9 & 76.4\\
			 MIL + Max + Gaussian & 86.0 & 72.1 & 83.1 & 78.5 & 74.5 & 76.3 & 83.2 & 59.3 & 56.7 & 68.4\\
			 MIL + Top 20 & 86.7 & 78.3 & 86.6 & 86.9 & 79.6 & 75.2 & 86.5 & 64.0 & 63.8 & 56.8\\
			 Sparse coding (200)   & 86.9 & 48.6 & 80.6 & 81.0 & 82.8 & 57.4 & 82.8 & 71.7 & 56.1 & 72.2 \\
			 Sparse coding (4,000) & 93.6 & 74.5 & 89.8 & 86.7 & 94.5 & 86.1 & 92.8 & 78.7 & 76.8 & \textbf{86.0}\\
			 Ours & \textbf{95.4} & \textbf{82.2} & \textbf{91.2} & \textbf{93.0} & \textbf{94.6} & \textbf{92.8} & \textbf{95.1} & \textbf{92.8} & \textbf{92.0} & 74.8 \\
            \hline \noalign{\smallskip}
            Methods & horse & house & motorbike & road & shoes & sofa &street & table lamp & train & tree & \textbf{mAP} \\

            \noalign{\smallskip}

            \hline

            \noalign{\smallskip}
			 Positive-negative Ratio & 23.9 & 47.4 & 30.9 & 48.2 & 56.4 & 39.7 & 42.7 & 16.9 & 28.6 & 44.7 & 41.4 \\
             Global SVM & 73.3 & 82.0 & 75.6 & 81.3 & 88.2 & 77.7 & 73.8 & 66.5 & 69.2 & 73.9 & 77.3\\
             MIL + Max & 70.3 &	 80.0 & 74.8 & 78.1 & 87.7 & 76.4 & 69.1 & 65.1 & 67.3 & 77.0 & 76.3\\
			 MIL + Max + Gaussian & 63.1 & 74.6 & 65.9 & 66.1 & 85.8 & 69.7 & 55.5 & 60.5 & 64.1 & 69.8 & 70.7\\
			 MIL + Top 20 & 63.7 & 76.4 & 76.9 & 73.6 & 90.3 & 69.7 & 63.7 & 52.3 & 67.2 & 75.2 & 73.7\\
			 Sparse coding (200)   & 61.5 & 71.3 & 61.0 & 80.1 & 82.3 & 80.2 & 84.1 & 52.3 &
65.5 & 57.6 & 70.8\\
			 Sparse coding (4,000) & 80.0 & 89.3 & 75.5 & 89.9 & 87.2 & 87.7 & 91.1 & 67.9 & 81.9 & 78.9 & 84.4\\
			 Ours & \textbf{85.4} & \textbf{94.4} & \textbf{85.0} & \textbf{90.8} & \textbf{95.3} & \textbf{88.9} & \textbf{94.8} & \textbf{78.3} & \textbf{91.3} & \textbf{85.0} & \textbf{89.7}\\
            \hline\\

      \end{tabular}

  \label{tab:result}%
\end{table*}%

We compare our method to the following methods.

\noindent \textbf{Positive-negative Ratio} If we apply an object detector to the image regions, considerable portion of the regions of a regular image should be positively scored. While on the contrary, images of other classes are supposed to have negatively-scored proposals only. Based on this intuitive assumption, we use the ratio of positive proposal number to the number of negative proposals within one image as its representation to construct two Gaussian models for regular images and ``other class'' images separately. Given a test image, we determine whether it is irregular via evaluating its fitting degree to these two Gaussians.

\noindent \textbf{Global SVM} According to the analysis in \textbf{Postulate I} in Section \ref{sec:motivation}, the classification score of an image reflects the degree of containing the regular object-of-interest and the scores of the three types of images (regular, irregular, other class) should form the relationship of $f(I^o)<f(I^u)<f(I^r)$. For this method, we train a classifier for each class based on the global features of the images using linear SVM \cite{Fan:2008} where regular images are used as positive data and ``other class'' images are treated as negative data.
Assuming the mean of the decision scores of irregular images is 0, we use negative absolute value of the decision score $-|f(I^t)|$ as the irregularity measurement for a test image $I^t$.

\noindent \textbf{MIL + Max} The global representation of an image is a mixture of the patterns of both the object-of-interest and the background. To avoid the distraction influence of the background, for the second solution we use the maximum proposal-level score $f_{max}(I^t)$ as the decision score of each image based on the object detector learned from MIL. Similarly we use $-|f_{max}(I^t)|$ as the irregularity measurement.

\noindent \textbf{MIL + Max + Gaussian} Different from above \textbf{MIL + Max} strategy, we take into consideration the uncertainty of the distribution of the maximum detection scores via modelling the maximum scores of regular images $\mathbf{I^r}$ and ``other class'' images $\mathbf{I^o}$ using two Gaussian distributions separately. We use maximum likelihood to estimate the parameters of these two Gaussians (means and variances). Given a test image $I^t$,
we can calculate the likelihood of the image belonging to regular images as
$p(I^t|\mathbf{I^r})$ and similarly the possibility of belonging to other classes as $p(I^t|\mathbf{I^u})$. Since an irregular image is expected to be able to fit neither of these two models, we set the final score of a test image as $-\mathrm{max}(p(I^t|\mathbf{I^r}), p(I^t|\mathbf{I^u}))$.

\noindent \textbf{MIL + Top k} Instead of using the maximum score only, for this method, we obtain the image-level score $f_{topk}(I^t)$ of a test image $I^t$ by averaging the top $k$ scores of its proposals. And the final score for an image is $-|f_{topk}(I^t)|$.

\noindent \textbf{Sparse coding} Similar to \cite{BinZhao:2011}, we use sparse coding based reconstruction error as the criterion for irregular image detection. The assumption is that both regular images and ``other class'' images can be well reconstructed by their corresponding dictionaries. For each class, we learn dictionaries for regular images and ``other class'' images separately. We try dictionary size 200 (\cite{BinZhao:2011} uses 200), 4,000 and 5,000. Given a test image $I^t$, we infer the coding vectors of its proposals and calculate the reconstruction residues of the proposals. Let $r^t_{r}$ be the mean residue for this image calculated based on the dictionary learned from regular images and $r^t_{o}$ be the mean residue based on the dictionary learned from ``other class'' images. The irregularity measurement for a test image can be calculated as $\mathrm{min}(r^t_{r}, r^t_{o})$.

\subsubsection{Quantitative Results}
\begin{figure*}[ht]
\begin{center}
\includegraphics[scale=0.325]{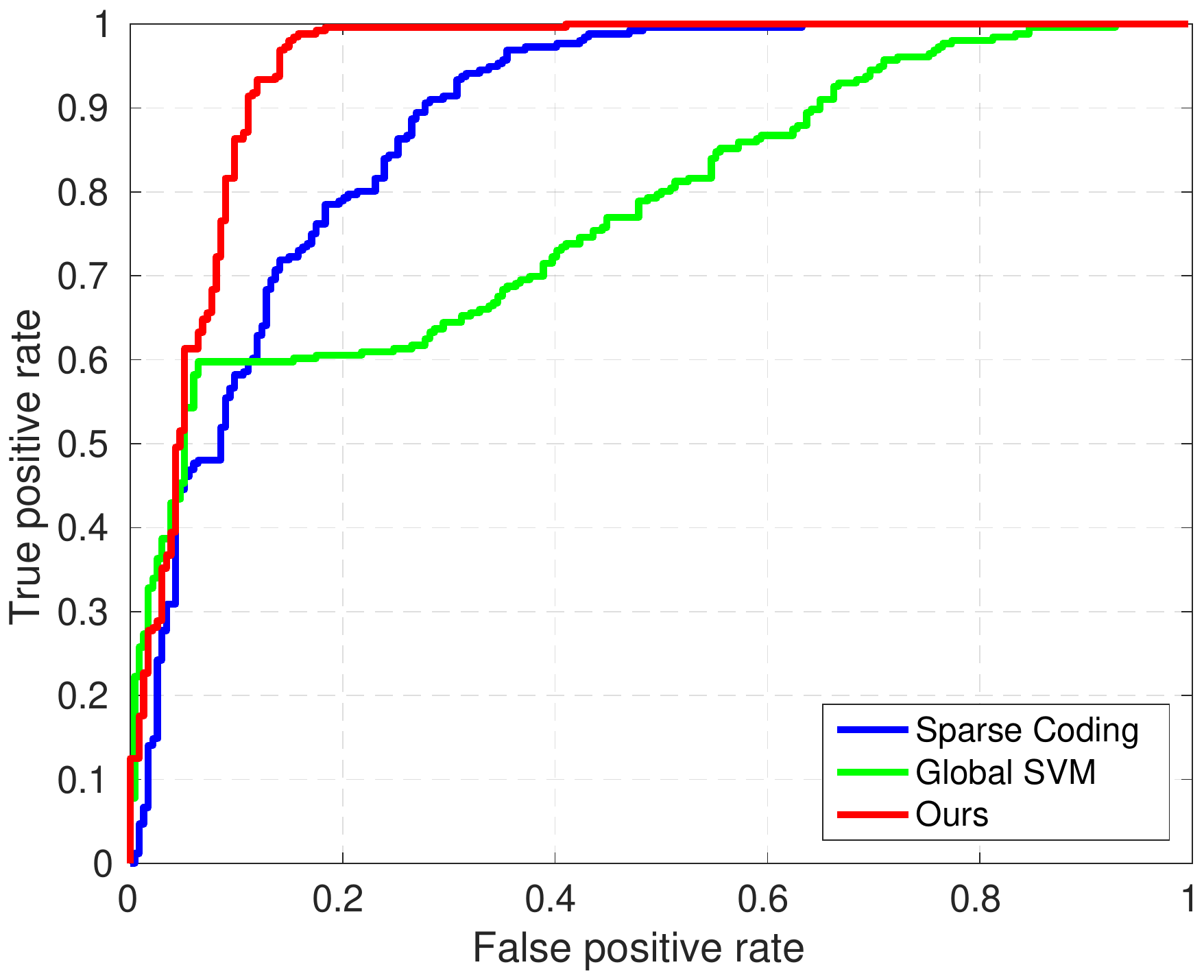}
\includegraphics[scale=0.325]{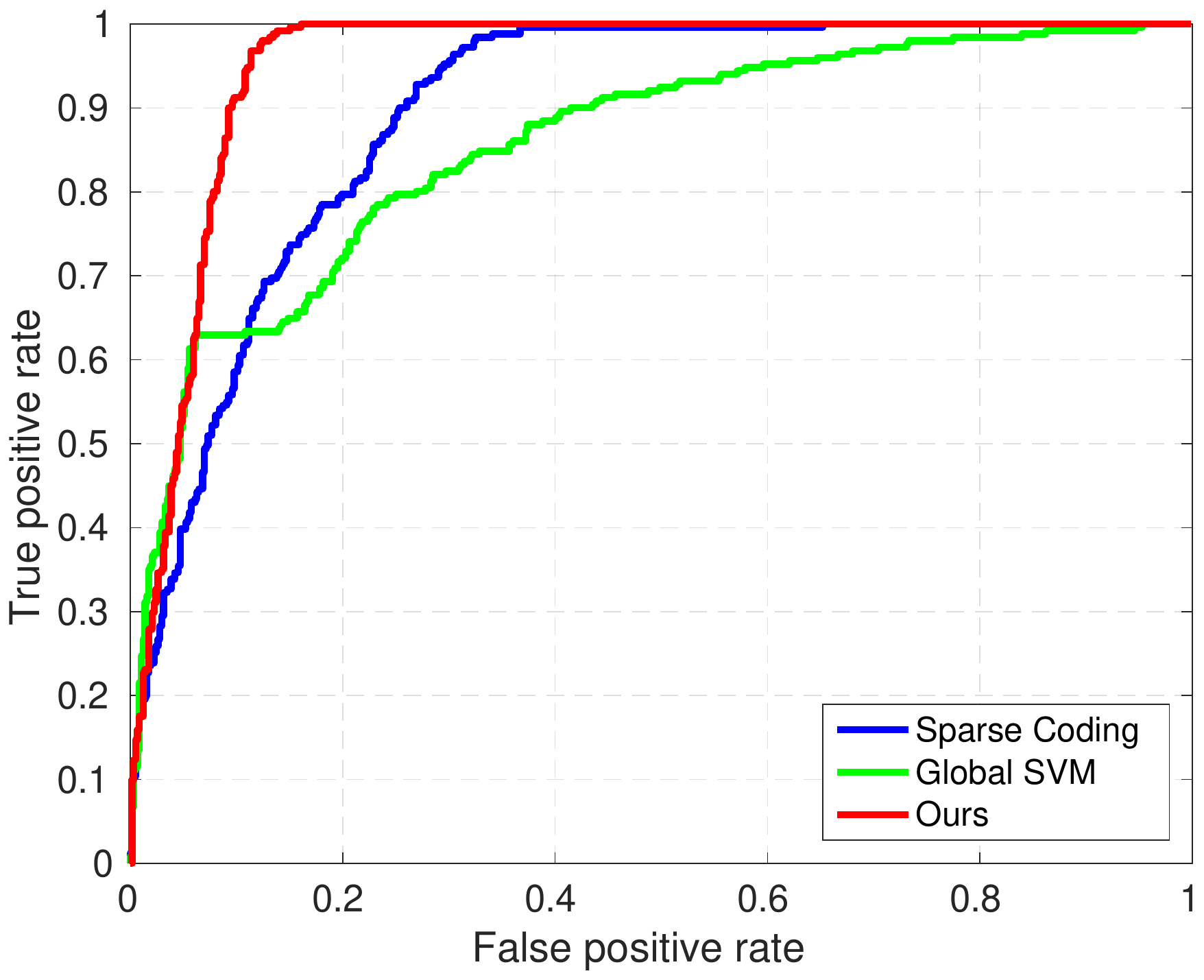}
\includegraphics[scale=0.325]{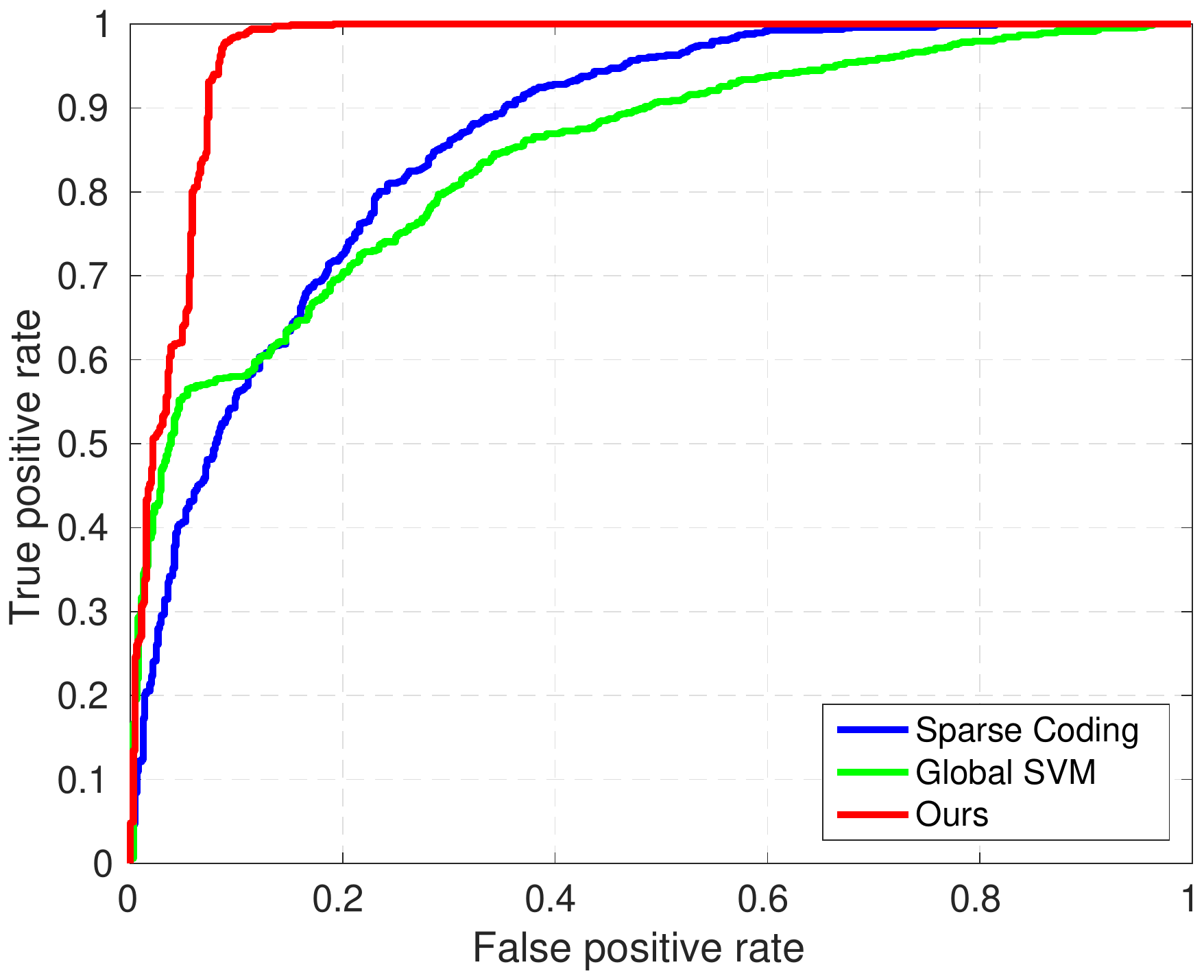}
\end{center}
\caption{ROC curve for \textbf{Sparse coding, Global SVM} and \textbf{our method} on three categories. From left to right: boat, motorbike, shoes.}
\label{fig:roc}
\end{figure*}

\begin{figure*}
\begin{center}
\includegraphics[scale=0.45]{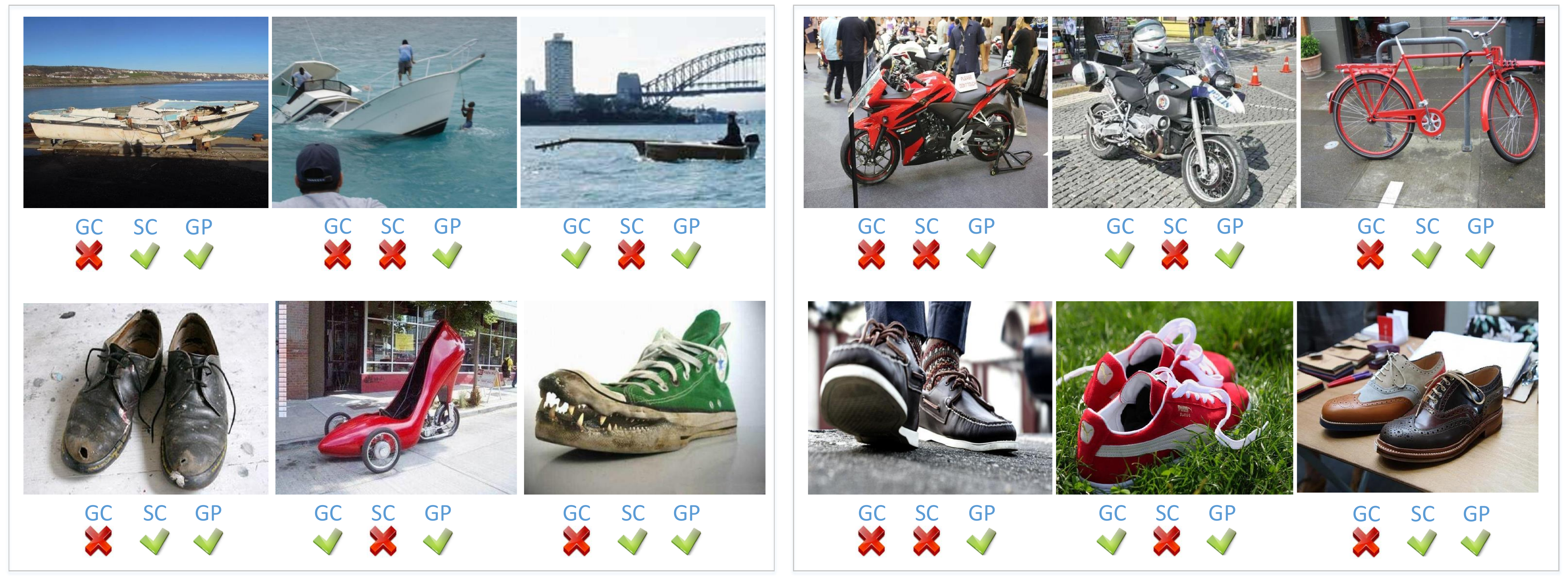}
\end{center}
\caption{Qualitative performance comparison between our method (GP) and two alternative solutions, \textbf{Global SVM} (GC) and \textbf{Sparse coding} (SC). Left column displays the \textbf{false negative} examples when fixing the false positive rate to be 0.2 where cross mark indicates false negative and check mark indicates true positive. Right column displays the \textbf{false positive} examples when fixing the true positive rate to be 0.9 where cross mark denotes false positive and check mark denotes true negative. Three categories are boat, shoes and motorbike.}
\label{fig:resultVis}
\end{figure*}

Table \ref{tab:result} shows the quantitative results. As can be seen, our method outperforms other compared methods. Also we show the ROC performances of our method and two most competitive methods on some example categories in Fig.~\ref{fig:roc}. Both these two measurements demonstrate the effectiveness of the proposed method.

The proposal ratio based method performs worst among these methods which indicates that the irregularity detection cannot be achieved by simply counting the number of positive and/or negative proposals. There are two reasons. The first is that the number of proposals varies between different images and the second reason is that for some irregular object images e.g., images of severely damaged cars, there may be no positively scored proposals detected.

The next four methods are classification-based methods. While the first three use single score per image from either the global image or the region with maximum detection score, \textbf{MIL+Top k} utilizes multiple region scores but treat them as i.i.d. \textbf{Global SVM} achieves a mAP of $77.3\%$ (when using fully-connected features only, we obtain $75.4\%$) which to some extent justifies \textbf{Postulate I}. However, as illustrated in Fig.~\ref{fig:hist}, this strategy fails to distinguish some irregular images
that obtain extreme high or low decision scores. A drawback of using image-level representation is that the background can influence the decision score especially when the background dominates the image. Multi-instance learning is supposed to be a remedy because it makes it possible to focus on the object-of-interest via considering the proposal with maximum detection score. But using maximum detection score alone may risk missing the irregular part of the object. From Table \ref{tab:result}, we can see \textbf{MIL+Max} obtains comparable results to \textbf{Global SVM}. To take into consideration the uncertainty of the detection scores, rather than directly using the maximum detection scores, we construct Gaussian models for the maximum scores of regular images and ``other class'' images separately and determine whether an image is irregular via evaluating its fitting likelihood to these two Gaussian models. However, the performance degrades to $70.7\%$. The reason may be that the distribution of the maximum detection scores is not strictly Gaussian. Instead of using the maximum detection score of each image, in \textbf{MIL+Top20}, we aggregate the top 20 scores of each image via average pooling. Benefiting from this strategy, the performances on some classes like \textsl{apple, boat} are obviously boosted. However, on some other classes such as \textsl{horse, table lamp} it shows inferior performance to \textbf{Global SVM} and \textbf{MIL+Max}. As can be seen, our method significantly outperforms this strategy on all the classes. This big gap may to a large extent result from our capabilities of modelling the inter-dependencies of the proposal-level scores within one image.

For sparse coding, we first test the performance using dictionaries of size 200 as \cite{BinZhao:2011} and the result is unsatisfactory which means 200 bases are not sufficient to cover the feature spaces of regular images or ``other class'' images. When the dictionary size is increased to 4,000, the performance is significantly improved. But after that continuing to increase the dictionary size (we test \textbf{5,000}) can lead to no improvement any more. Our method outperforms sparse coding by $5.3\%$. Apart from effectiveness, our method is also more efficient than sparse coding. Given a test image, while sparse coding needs to infer the coding vector for the high-dimensional appearance features our method works on quite low-dimensional space as defined in Eq.~(\ref{eq:inter-cov}).

\subsubsection{Qualitative Results}
Fig.~\ref{fig:resultVis} demonstrates the qualitative comparison between our method and two compared methods \textbf{Global SVM (GC)} and \textbf{Sparse coding (SC)} on three object categories that are boat, motorbike and shoes. Comparing to our method, GC suffers from two drawbacks: 1) it subjects to the distraction influence of the background, and 2) it may ignore the fine details of the objects. Due to the influence of the background, GC may mistakenly classify
the regular object within complex background into irregular object like the ``shoes'' on the right side of Fig.~\ref{fig:resultVis}. Also, only looking at the global appearance makes it hard for GC to identify some irregular objects with fine irregularities such as the ``broken boat'' and ``broken shoes'' in Fig.~\ref{fig:resultVis}. SC has similar deficiency that is it can be distracted or even dominated by the background. For example, the ``capsized boat'' is identified as ``regular boat'' while ``regular motorbike'' within complex background is regarded as ``irregular motorbike''. Comparing to these two methods our method is more robust. While using detection scores enables us to getting rid of the distraction influence of the background, modelling the inter-dependencies of the detection scores at multiple regions can help us to effectively discover the finer irregularities.

\section{Conclusions}
We have proposed a novel approach for the task of irregular object identification in an ``open world''
setting via inspecting the detection score patterns of an image.
We have proposed to use Gaussian Processes to model the values as well the spatial distribution of the detection scores.
Our method shows superior performance against some compared methods on a large dataset presented in this work.

{
\bibliographystyle{ieee}
\bibliography{cvml,CSRef}
}

\end{document}